\documentclass[letterpaper, 10 pt, conference]{ieeeconf}  

\IEEEoverridecommandlockouts                              

\usepackage{cite}
\usepackage{amsmath,amssymb,amsfonts}
\usepackage{algorithmic}
\usepackage{graphicx}
\usepackage{color,calc}
\usepackage{makecell}
\usepackage{booktabs}
\usepackage{array}
\usepackage{ragged2e}
\usepackage{caption}
\usepackage{multirow}
\usepackage{url}
\usepackage{listings}
\usepackage{xcolor}
\usepackage{subcaption}
\usepackage{cleveref}
\usepackage[ruled,vlined,linesnumbered]{algorithm2e}
\usepackage{comment}
\usepackage{tikz}
\usetikzlibrary{positioning, arrows.meta, shapes.geometric, fit, backgrounds, shadows}

\lstdefinelanguage{ROSsrv}{
  morekeywords={string,---},
  sensitive=false,
  morecomment=[l]{\#},
}

\lstdefinelanguage{_common}{
  sensitive=false,
  morecomment=[l]{\#},
}

\title{\LARGE \bf
From USD Scenes to Knowledge Graphs: Zero-Shot Ontology Grounding with LLMs 
}

\author{Jiangtao Shuai$^{1}$, Zongxiong Chen$^{2}$, Manfred Hauswirth$^{1, 2}$, Sonja Schimmler$^{1, 2}$
\thanks{*This research has received funding from the German Research Foundation (DFG) under project numbers
441926934 (NFDI4Cat) and 460234259 (NFDI4DataScience)}
\thanks{$^{1}$The authors are with 
        Technical University of Berlin, Germany and $^{2}$Fraunhofer FOKUS, Germany.
        {\tt\small \{firstname.lastname\}@tu-berlin.de}; {\tt\small \{firstname.lastname\}@fokus.fraunhofer.de}}%
}
\begin{document}
\maketitle
\thispagestyle{empty}
\pagestyle{empty}

\begin{abstract}
Constructing knowledge graphs from 3D simulation scenes is essential 
for robot task reasoning, but the key bottleneck, grounding scene 
objects to formal ontology classes, still relies on manually curated 
dictionaries that are brittle and do not generalize across assets. We 
investigate whether large language models (LLMs) can automate this 
grounding step for Universal Scene Description (USD) scenes as a 
zero-shot, training-free alternative. On a kitchen scene (125 objects) with SOMA-HOME Ontology, 
LLMs achieve 90--96\% exact-match accuracy with 
descriptive names and 49--89\% with abbreviated names, substantially 
outperforming dictionary and embedding baselines. Under fully opaque 
names, context-augmented prompting recovers up to 48\%. Feature 
ablation reveals that LLMs primarily exploit semantic cues in the 
scene graph (sibling names and parent paths); anonymizing these cues 
reduces accuracy to 0--6\%, while geometry alone yields only 4--17\%.
\end{abstract}

\section{Introduction}
\label{sec:intro}

For autonomous robots, understanding objects as semantically grounded
entities is essential for knowledge-driven reasoning and task
execution~\cite{beetz2025robot}. A well-established approach is to 
represent environment knowledge using ontologies, which provide a formal 
schema for constructing knowledge graphs. Perceived objects are reflected in this schema as typed instances, enabling commonsense reasoning and 
task planning~\cite{beetz2018knowrob, bessler2021soma}.

Digital twins have emerged as a key infrastructure for robot software development and evaluation, 
yet their semantic potential remains largely untapped: 
scene elements are rarely linked to formal ontologies, and knowledge 
graphs, when present, serve as metadata stores rather than active 
reasoning components~\cite{karabulut2024ontologies}. Universal Scene 
Description (USD), increasingly adopted as the scene representation 
backbone in digital twin platforms, exemplifies this gap. USD 
encodes scenes as hierarchical graphs of primitives (\emph{prims}) 
with rich geometric attributes and parent-child relationships, but 
these structures reflect spatial organization rather than ontological 
semantics. Moreover, as a format designed for asset exchange across 
diverse users, USD relies on user-defined identifiers whose naming 
conventions vary widely, making consistent mapping to ontology classes 
nontrivial.

Prior work~\cite{nguyen2024translating, nguyen2025generating} 
constructs knowledge graphs from USD scenes by mapping prims to OWL 
classes (grounding) and instantiating ABox individuals. However, the 
grounding step relies on handcrafted dictionaries that require expert 
curation and do not generalize to diverse assets, preventing the 
pipeline from scaling without manual effort.

We present, to the best of our knowledge, the first study exploring 
whether LLMs can automate this grounding step, enabling zero-shot 
knowledge graph construction from USD scenes. Specifically, 
we ask: \emph{Can an LLM infer the correct ontology class of a USD prim solely 
from information available in the USD scene description, and what 
roles do different information sources play in this grounding process?}

\begin{figure}[tb]
    \centering
    \includegraphics[width=0.95\linewidth]{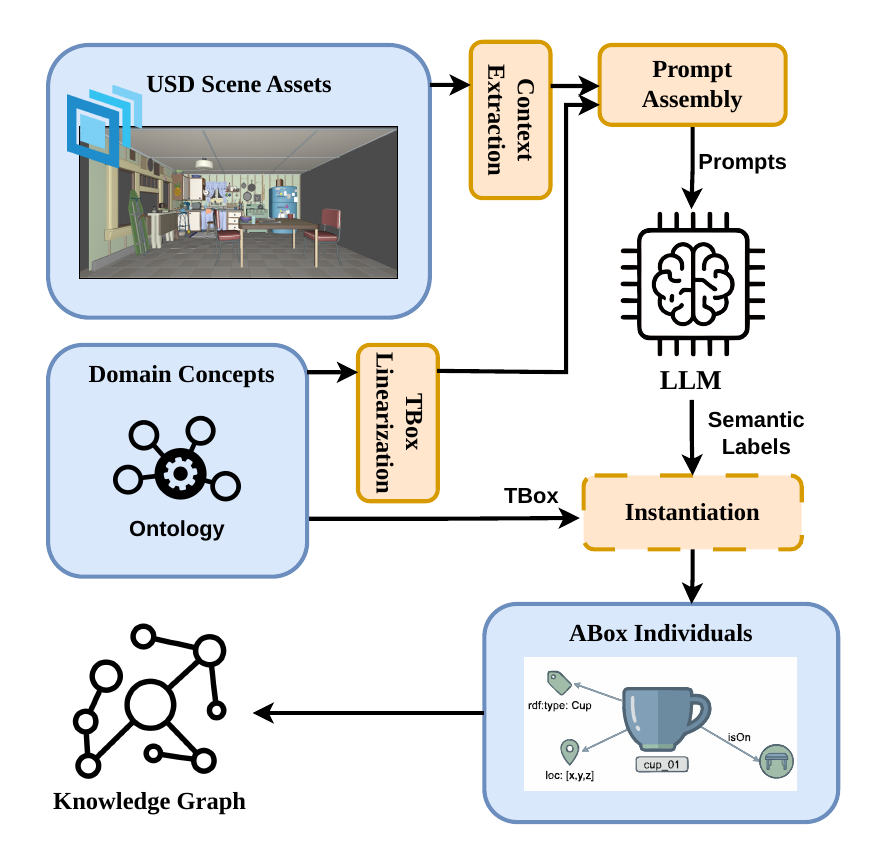}
    \caption{Overview of the proposed LLM-based ontology grounding 
pipeline for USD scenes. 
USD scene attributes are extracted and 
combined with a linearized ontology TBox into prompts; the LLM 
predicts semantic labels (ontology classes) for each prim, which are 
then instantiated as ABox individuals to form a knowledge graph.
Blue boxes denote data sources and 
artifacts; orange boxes denote processing steps; dashed borders 
indicate modules adopted from the Multiverse 
framework~\cite{nguyen2024translating}. The rendered kitchen scene is adopted 
from~\cite{openusd}.}

    \label{fig:overview}
    \vspace{-0.4cm}
\end{figure}

Our contributions are summarized as follows:
\begin{itemize}
\item We present the first study of LLM-based ontology grounding 
for USD scenes, showing that off-the-shelf LLMs provide a zero-shot 
alternative to handcrafted mapping for semantic grounding in knowledge-graph construction.
\item We empirically evaluate LLM-based ontology grounding under varying naming conditions, 
demonstrating strong performance with descriptive identifiers and reasonable robustness under reduced lexical cues.
\item We provide an in-depth analysis of the factors driving ontology grounding performance, revealing that ontology class assignment is driven primarily by semantic cues embedded in USD scene graphs.
\end{itemize}


\section{Problem Formulation}
\label{sec:problem}

Constructing a knowledge graph from a USD scene requires two steps:
(1)~\emph{grounding}---mapping each prim to an ontology class, and
(2)~\emph{instantiation}---creating ABox~\cite{funk2025fitting} individuals with spatial and relational properties derived from the scene graph.
Instantiation can be automated given correct class assignments (e.g., using tools from~\cite{nguyen2024translating}); therefore, we focus on the grounding problem.
To formalize this question, we have to specify (i)~what information 
the grounding function $g$ has available, and (ii)~how the 
contribution of each information source is measured. We address~(i) below and~(ii) 
through the experimental design in \Cref{sec:experiments}.

Formally, let the set of prims in a USD scene be $\mathcal{S}=\{p_1,\dots,p_n\}$ and
let $\mathcal{C}$ denote the set of classes in an OWL TBox. Grounding is a mapping $g:\mathcal{S}\to\mathcal{C}$
that assigns each prim $p_i$ a class $c_i=g(p_i)\in\mathcal{C}$. 
When no leaf class provides a sufficient match, we instruct the model to return the most specific applicable superclass rather than an "unknown" label, 
so that downstream tasks can still leverage partial type information (e.g., knowing an object is \texttt{Furniture} even when its exact subclass is uncertain).


Each prim provides multiple information sources, including a user-defined name, geometric attributes (bounding box, mass), world-frame position, and scene-graph context (parent path and siblings). 
To study their individual contributions, we define three \emph{naming regimes} that progressively degrade the most accessible cue, the prim name: \emph{semantic}, \emph{abbreviated}, and \emph{opaque}. (\Cref{sec:setup})
Under opaque naming, grounding relies entirely on non-lexical features, directly testing the sufficiency of geometric and hierarchical context. 
We further perform a feature ablation (\Cref{sec:ablation}) over geometry, position, and hierarchy to quantify their individual contributions.

\section{Approach}
\label{sec:approach}

%
%
As illustrated in \Cref{fig:overview}, 
we convert structured USD attributes (Sec.~\ref{sec:context}) into natural-language prompts (Sec.~\ref{sec:prompt}), combined with a linearized ontology TBox (Sec.~\ref{sec:box_lin}). 
The LLM performs zero-shot grounding by selecting the best-matching class using pretrained world knowledge, as the TBox provides descriptions but not canonical object dimensions, e.g., linking a 1.9m bounding box to the class \texttt{Refrigerator} relies on the model's internalized priors about real-world object sizes.
To study the contribution of different signals, we design three prompting strategies that progressively incorporate USD-derived context (\Cref{sec:prompt}).

\subsection{Context Extraction from USD}
\label{sec:context}

For each prim $p_i$ in the USD stage, we extract three groups of features: 
(i) \textit{Geometry} (axis-aligned bounding box $(w, h, d)$ in meters and mass $m$ in kilograms, when available via \texttt{UsdPhysicsMassAPI}); 
(ii) \textit{Position} (world-frame translation $(x, y, z)$); and 
(iii) \textit{Hierarchy} (the prim's full parent path, e.g., \texttt{/.../SinkArea\_grp/Sink\_grp}, and the \emph{names} of its sibling prims in the scene graph).

These features are formatted as key-value pairs in plain text and
concatenated into the prompt (see supplementary material for an example).
Note that the hierarchy features carry two distinct types of information:
\emph{structural} information (depth, sibling count) and
\emph{lexical} information (semantic content of sibling and parent names).
As our ablation study will show (\Cref{sec:ablation}), the latter dominates.

\subsection{TBox Linearization}
\label{sec:box_lin}

The ontology TBox is linearized into a flat list of all relevant classes
(both leaf and intermediate nodes) with their parent class and
natural-language descriptions derived from \texttt{rdfs:comment} annotations.
For instance:
\begin{quote}\small\ttfamily
Bowl (subClassOf: Crockery) -- A round, deep object used to contain food or liquid.
\end{quote}
Including intermediate classes (e.g., \texttt{Crockery}, \texttt{DesignedContainer})
allows the model to select an appropriate superclass when no leaf class
provides a sufficient match.
The list is presented in alphabetical order; we do not preserve the tree
structure in the prompt, though positional effects in long class lists
may influence predictions~\cite{liu2024lost}.

\subsection{Prompting Strategies}
\label{sec:prompt}

    
    
We design three prompting strategies with increasing input richness. 
\textbf{(A) Name-only} uses only the prim name and the linearized TBox, isolating lexical cues. 
\textbf{(B) Context-augmented} additionally incorporates geometric, positional, and hierarchical context (\Cref{sec:context}), instructing the model to use all features and prefer parent classes when evidence is insufficient. 
\textbf{(C) Chain-of-thought (CoT)} uses the same inputs as (B) but requires step-by-step reasoning over geometry, mass, position, and scene-graph context before producing the final JSON mapping, encouraging explicit multi-feature integration at the cost of increased output tokens.
Notably, for Qwen in instruct mode (Q-NT), CoT does not trigger 
internal reasoning during inference; instead, it explicitly instructs 
the model to generate visible reasoning text 
(e.g., analyzing bounding-box dimensions, scene-graph context, and 
candidate classes) before committing to a classification.
\section{Experiments}
\label{sec:experiments}

\subsection{Experiment Setup}
\label{sec:setup}


We use the \textit{Kitchen Set} USD asset from Pixar~\cite{openusd} as our testbed, as it provides a high-quality and structurally rich scene suitable for evaluating grounding methods.
We construct a dataset of 125 object-level prims from approximately 2,700 scene prims by excluding grouping nodes, camera/light prims, sub-components (e.g., screws), and prims without physical geometry, enabling manual verification of ground-truth labels. 
The selected objects span diverse categories, including furniture, appliances, containers, cutlery, tools, and decorations, ensuring representativeness.


The target ontology is \textit{SOMA-HOME}~\cite{bessler2021soma}, a home-environment extension of SOMA with 94 kitchen-relevant classes organized in a multi-level hierarchy (max depth~5). 
Its functional organization (e.g., \texttt{Crockery}) creates semantically meaningful ambiguity among geometrically similar objects (e.g., cups vs.\ bowls), while class descriptions (\texttt{rdfs:comment}) provide natural-language cues without specifying canonical dimensions, requiring implicit world knowledge. 
As a publicly available ontology integrated into robotic systems such as KnowRob~\cite{beetz2018knowrob}, it enables our
results to be situated within an existing deployment pipeline. 
Of the 125 prims, 86 map to leaf classes directly, while 39 require superclass mapping (e.g., \texttt{Toaster}$\to$\texttt{Appliance}).

To construct naming regimes, we transform each prim name as follows: \emph{semantic} retains the 
original identifier (e.g., \texttt{CupCBlue\_1}); 
\emph{abbreviated} applies collision-free vowel removal with 
per-word truncation (e.g., \texttt{cpcbl1}); \emph{opaque} 
replaces the name with a sequential identifier 
(e.g., \texttt{obj\_042}).



We evaluate \textit{Gemini~2.5 Flash} (2.5F), \textit{Gemini~3 Flash Preview} (3F), and \textit{Qwen~3.5-27B}~\cite{yang2025qwen3} (Qwen), with Qwen tested in thinking (Q-T) and instruct (Q-NT) modes. 
All models use temperature~0 for deterministic decoding; detailed configurations can be found in the replication package~\url{https://github.com/JTShuai/USD_2_KG}

\subsection{Evaluation Metrics}
\label{sec:metric}

\paragraph{Exact-Match Accuracy (EMA)}
The fraction of prims that were assigned their ground-truth class:
\begin{equation}
    \text{EMA} = \frac{1}{n} \sum_{i=1}^{n} \mathbb{1}[\hat{c}_i = c_i^*]
\end{equation}
where $\hat{c}_i$ is the predicted class and $c_i^*$ the ground-truth class.

\paragraph{Hierarchical F1 (HF1)}
Since ontology classes form a tree, predictions that select ancestors or siblings of the ground-truth class are more informative than random guesses.
Following Silla and Freitas~\cite{silla2011hierarchical}, each class is expanded into its \emph{ancestor set} $\uparrow\!c = \{c, \text{parent}(c), \ldots, \text{root}\}$, and hierarchical precision ($hP$) and recall ($hR$) are defined as:
\begin{equation}
    hP = \frac{|\uparrow\!\hat{c}\;\cap\;\uparrow\!c^*|}{|\uparrow\!\hat{c}|},\quad
    hR = \frac{|\uparrow\!\hat{c}\;\cap\;\uparrow\!c^*|}{|\uparrow\!c^*|}
\end{equation}
The final score is the macro-averaged harmonic mean of $hP$ and $hR$:
\begin{equation}
    \mathrm{HF1} = \frac{1}{n}\sum_{i=1}^{n}\frac{2\,hP_i\,hR_i}{hP_i + hR_i}
\end{equation}
This metric assigns full credit to exact matches and partial credit to predictions that are close in the class hierarchy.

\subsection{Baselines}
\label{sec:baseline}
We compare against three non-LLM baselines:
\begin{itemize}
    \item \emph{Multiverse (Dictionary Mapping)}: the dictionary-based mapping
    from~\cite{nguyen2024translating}, which maps prim names to OWL
    classes via exact string lookup. 
    \item \emph{S-BERT}: cosine similarity between
    Sentence-BERT~\cite{sbert2019} embeddings
    (\texttt{all-MiniLM-L6-v2}) of the prim name and each ontology
    class name; the nearest neighbor is selected.
    \item \emph{Log-Volume Nearest Neighbor (LV-NN)}: each ontology class is assigned an
    approximate bounding-box volume based on common-sense priors
    (e.g., Refrigerator $\approx$ 1.5\,m$^3$,
    Cup $\approx$ 0.5$\times 10^{-3}$\,m$^3$).
    Each prim is matched to the class with the nearest volume on a
    log scale. This baseline is name-independent and uses only
    geometric information.
\end{itemize}
Note that supervised baselines are infeasible in our single-scene setting
as all labeled data serves as the test set.

\subsection{Main Results}

\Cref{tab:main} presents results across all methods and naming regimes.

\begin{table}[t]
\centering
\caption{Grounding performance on the Pixar Kitchen Set (125 prims) using the SOMA-HOME ontology (94 classes) under three naming regimes (semantic, abbreviated, opaque).
Results are reported as exact-match accuracy (EMA, \%) and hierarchical F1 (HF1). 
(A)--(C) denote prompting strategies with increasing input richness (see \Cref{sec:prompt}). 
Non-LLM baselines are shown at the top; 2.5F\,=\,Gemini~2.5 Flash, 3F\,=\,Gemini~3 Flash, Q-NT\,=\,Qwen~3.5-27B instruct, and Q-T\,=\,Qwen~3.5-27B thinking. 
``fail'' indicates degenerate outputs (see text).}
\label{tab:main}
\setlength{\tabcolsep}{2.5pt}
\footnotesize
\begin{tabular}{l cc cc cc}
\toprule
\multirow{2}{*}{\textbf{Method}} & \multicolumn{2}{c}{\textbf{Semantic}} & \multicolumn{2}{c}{\textbf{Abbreviated}} & \multicolumn{2}{c}{\textbf{Opaque}} \\
\cmidrule(lr){2-3} \cmidrule(lr){4-5} \cmidrule(lr){6-7}
 & EMA & HF1 & EMA & HF1 & EMA & HF1 \\
\midrule

Multiverse  &  9 & .09 &  0 & .00 &  0 & .00 \\
LV-NN          &   2 & .50 &  2  & .50 &  2  & .50 \\
S-BERT      &  67 & .85 & 14  & .52 &  2  & .53 \\
\midrule
2.5F (A)         &  \textbf{96} & \textbf{.99} & 84  & .93 &  6  & .50 \\
2.5F (B)         &  94 & .98 & 77  & .93 & 33  & .70 \\
2.5F (C)         &  94 & .98 & 76  & .91 & 30  & .70 \\
\midrule
3F (A)           &  94 & \textbf{.99} & 80  & .93 &  6  & .42 \\
3F (B)           &  92 & .98 & \textbf{89} & .96 & \textbf{48} & \textbf{.79} \\
3F (C)           &  94 & .98 & 83  & \textbf{.97} & \textbf{48} & .78 \\
\midrule
Q-NT (A)         &  94 & .98 & 67  & .83 &  6  & .50 \\
Q-NT (B)         &  90 & .98 & 54  & .79 & 17  & .62 \\
Q-NT (C)         &  90 & .98 & 49  & .68 & 26  & .68 \\
\midrule
Q-T (A)          &  94 & .98 & 75  & .90 &  fail & fail  \\
Q-T (B)          &  94 & \textbf{.99} & 69  & .77 & 19  & .50 \\
Q-T (C)          &  90 & .98 & 73  & .89 & 21  & .53 \\
\bottomrule
\end{tabular}
\end{table}

\paragraph{Can LLMs ground USD prims from names alone?}
All LLMs achieve 90--96\% EMA on semantic names and retain 49--89\% on 
abbreviated names, substantially outperforming S-BERT (67\%/14\%), the 
Multiverse (9\%/0\%), and LV-NN (2\% across all regimes).
The Multiverse's low score reflects its limited vocabulary; 
LV-NN's constant performance confirms that bounding-box volume alone 
cannot support fine-grained classification.
These results indicate that LLMs can reliably ground USD prims in a 
zero-shot setting when name cues are available.

\paragraph{Can geometric and hierarchical context compensate for 
missing names?}
With only the opaque name and TBox, Strategy~(A) scores 6\% across all 
models. 
Adding geometric and hierarchical context~(B) raises performance to 17--48\% (Gemini~3F achieves the highest (48\%)), 
confirming that non-lexical USD signals carry  grounding-relevant information
However, as the 
ablation in \Cref{sec:ablation} reveals, much of this gain stems from 
semantic sibling names and parent paths that remain visible in the 
opaque regime.

\paragraph{Chain-of-thought.}
CoT~(C) yields inconsistent gains over context-only~(B) under opaque naming: Q-NT improves by 9~points (17\%$\to$26\%), while other models show comparable or slightly degraded performance. 
No consistent benefits are observed under abbreviated or semantic naming. 
Notably, Q-NT is the only configuration without internal reasoning, which may explain its larger gains: prompt-induced reasoning serves as its primary mechanism for multi-feature integration, whereas models with built-in reasoning (Q-T, Gemini) benefit less from additional external guidance. 
These results suggest that explicit step-by-step reasoning may benefit models without internal reasoning capabilities under challenging conditions, but is not universally effective, indicating the need for more targeted CoT designs.

\paragraph{Thinking mode failure by Qwen 3.5-27B.}
Q-T~(A) on opaque names (marked ``fail'' in \Cref{tab:main})
produces no usable output.
The model's internal reasoning first enters a deliberation loop,
repeatedly reconsidering whether classification is possible
without semantic cues.
It then degenerates into emitting semantically vacuous tokens: the recurring fragment ``\emph{wheretogetherthroughwithwithou}'' grows progressively longer with each repetition, accumulating meaningless concatenations until the output budget is exhausted. 
This failure mode illustrates that, under extended reasoning without grounding signals, Qwen~3.5-27B can collapse into self-reinforcing token repetition.

\subsection{Feature Ablation}
\label{sec:ablation}
To determine what roles different USD-derived signals play in 
grounding, we run Strategy~(B) on opaque names providing only a 
single feature subset at a time. \Cref{tab:ablation} reports EMA 
for each condition.

\begin{table}[t]
\centering
\caption{Feature ablation of context under opaque naming using Strategy~(B) (125 prims, 94 classes). 
Each row corresponds to a subset of input features provided to the LLM. 
``anon.'' denotes conditions where semantic sibling names and parent paths are replaced with opaque identifiers. 
Results are reported as exact-match accuracy (EMA, \%).}
\label{tab:ablation}
\setlength{\tabcolsep}{3pt}
\small
\begin{tabular}{@{}l cccc@{}}
\toprule
\textbf{Condition} & \textbf{2.5F} & \textbf{3F} & \textbf{Q-NT} & \textbf{Q-T} \\
\midrule
Geometry only       & 17 & 17 & 11 &  4 \\
Hierarchy only      & 28 & \textbf{54} & 21 & 22 \\
Geo + position      & 25 & 12 &  6 &  9 \\
Position only       &  4 &  7 &  6 &  9 \\
\midrule
Full ctx.\ (anon.)  & 18 & 15 & 25 &  8 \\
Hier.\ only (anon.) &  0 &  6 &  6 &  5 \\
\bottomrule
\end{tabular}
\end{table}

Hierarchy (parent paths + sibling names) is the strongest single 
feature across all models (21--54\%), while geometry alone scores 
4--17\% and position alone 4--9\%. Notably, Gemini~3F achieves 54\% 
with hierarchy alone, exceeding its full-context opaque~(B) score 
(48\% in \Cref{tab:main}), suggesting that geometric and positional 
features can introduce noise.

The anonymization control (bottom rows) pinpoints the source: when 
semantic sibling names and parent paths are replaced with opaque 
identifiers, hierarchy-only EMA drops to 0--6\% and full-context EMA 
to 8--25\%. 
This confirms that the LLM exploits contextual semantic cues in the 
scene graph rather than geometric reasoning to infer class assignments,
even when the target prim's own name is uninformative.
While prim names remain the most effective signal for grounding, 
the scene-graph context alone enables meaningful classification 
even when names are entirely absent.

\subsection{Error Analysis}
On opaque~(B), Gemini~3F leads with 48\% (60/125).
Objects correctly identified across models tend to have
distinctive bounding-box dimensions
(e.g., the 5.8\,m-wide room, 1.6\,m-tall stove, 1.9\,m-tall refrigerator)
or unique geometric profiles (e.g., the elongated broom at 0.4$\times$0.7$\times$1.4\,m).
The most common failure mode is \emph{superclass collapse}:
geometrically similar objects (e.g., cups, bowls, flower pots, small bottles) are
mapped to shallow superclasses such as \texttt{Crockery} or \texttt{DesignedContainer}.
Among the 39 prims whose ground truth is already a superclass
(e.g., \texttt{FlowerPot}$\to$\texttt{DesignedContainer}),
models frequently predict a sibling subclass (e.g., \texttt{Cup}),
confirming that these objects are inherently ambiguous
when only geometric features are available.
\section{Discussion and Conclusion}
\label{sec:discussion}

\textbf{Limitations.}
Our evaluation covers a single kitchen scene with 125 objects; generalization to other domains (e.g., warehouse, office) and larger-scale scenes remains to be validated.
All models are evaluated with temperature~0 in a single deterministic run, so we do not report variance estimates.
The opaque naming regime anonymizes only the target prim's name while retaining semantic sibling names and parent paths; a fully anonymized scene would better isolate geometric reasoning, though our anonymization control (\Cref{sec:ablation}) partially addresses this.
Finally, we linearize the TBox as a flat class list; preserving the ontology hierarchy in the prompt may improve superclass selection but increase prompt length.

\textbf{Conclusion.}
We investigated whether LLMs can automate ontology grounding for USD 
scenes, replacing handcrafted dictionaries with a zero-shot, 
training-free approach. On descriptive names, LLMs achieve 90--96\% 
accuracy; under fully opaque naming, context-augmented prompting 
recovers up to 48\% by exploiting scene-graph context. Feature 
ablation shows that this performance is primarily driven by semantic 
cues in sibling names and parent paths: anonymizing these cues reduces 
accuracy to 0--6\%, whereas geometry alone yields 4--17\%. For 
well-named assets, LLM-based grounding can directly replace 
dictionary-based pipelines; for degraded names, it provides a useful 
prior when scene-graph context is available. Future work includes 
multi-modal grounding with rendered thumbnails and integration into 
existing pipelines with human-in-the-loop confirmation for 
low-confidence predictions.

\bibliographystyle{IEEEtran}
\bibliography{./references}

\begin{thebibliography}{10}
\providecommand{\url}[1]{#1}
\csname url@samestyle\endcsname
\providecommand{\newblock}{\relax}
\providecommand{\bibinfo}[2]{#2}
\providecommand{\BIBentrySTDinterwordspacing}{\spaceskip=0pt\relax}
\providecommand{\BIBentryALTinterwordstretchfactor}{4}
\providecommand{\BIBentryALTinterwordspacing}{\spaceskip=\fontdimen2\font plus
\BIBentryALTinterwordstretchfactor\fontdimen3\font minus \fontdimen4\font\relax}
\providecommand{\BIBforeignlanguage}[2]{{%
\expandafter\ifx\csname l@#1\endcsname\relax
\typeout{** WARNING: IEEEtran.bst: No hyphenation pattern has been}%
\typeout{** loaded for the language `#1'. Using the pattern for}%
\typeout{** the default language instead.}%
\else
\language=\csname l@#1\endcsname
\fi
#2}}
\providecommand{\BIBdecl}{\relax}
\BIBdecl

\bibitem{beetz2025robot}
M.~Beetz, G.~Kazhoyan, and D.~Vernon, ``Robot manipulation in everyday activities with the cram 2.0 cognitive architecture and generalized action plans,'' \emph{Cognitive Systems Research}, vol.~92, p. 101375, 2025.

\bibitem{beetz2018knowrob}
M.~Beetz, D.~Be{\ss}ler, A.~Haidu, M.~Pomarlan, A.~K. Bozcuoglu, and G.~Bartels, ``{KnowRob} 2.0 -- a 2nd generation knowledge processing framework for cognition-enabled robotic agents,'' in \emph{Proc. IEEE Int. Conf. Robotics and Automation (ICRA)}, 2018, pp. 512--519.

\bibitem{bessler2021soma}
D.~Be{\ss}ler, R.~Porzel, M.~Pomarlan, A.~Vyas, S.~H{\"o}ffner, M.~Beetz, R.~Malaka, and J.~Bateman, ``Foundations of the socio-physical model of activities ({SOMA}) for autonomous robotic agents,'' in \emph{Proc. 12th Int. Conf. Formal Ontology in Information Systems (FOIS)}.\hskip 1em plus 0.5em minus 0.4em\relax IOS Press, 2021, pp. 159--174.

\bibitem{karabulut2024ontologies}
E.~Karabulut, S.~F. Pileggi, P.~Groth, and V.~Degeler, ``Ontologies in digital twins: A systematic literature review,'' \emph{Future Generation Computer Systems}, vol. 153, pp. 442--456, 2024.

\bibitem{nguyen2024translating}
G.~H. Nguyen, D.~Be{\ss}ler, S.~Stelter, M.~Pomarlan, and M.~Beetz, ``Translating universal scene descriptions into knowledge graphs for robotic environment,'' in \emph{2024 IEEE International Conference on Robotics and Automation (ICRA)}.\hskip 1em plus 0.5em minus 0.4em\relax IEEE, 2024, pp. 9389--9395.

\bibitem{nguyen2025generating}
G.~Nguyen, M.~Pomarlan, S.~Jongebloed, N.~Leusmann, M.~N. Vu, and M.~Beetz, ``Generating actionable robot knowledge bases by combining 3d scene graphs with robot ontologies,'' in \emph{2025 IEEE/RSJ International Conference on Intelligent Robots and Systems (IROS)}.\hskip 1em plus 0.5em minus 0.4em\relax IEEE, 2025, pp. 21\,527--21\,534.

\bibitem{openusd}
{Pixar Animation Studios}, ``Universal scene description (usd) documentation,'' \url{https://openusd.org/release/index.html}, accessed: 2026-04-09.

\bibitem{funk2025fitting}
M.~Funk, M.~Grosser, and C.~Lutz, ``Fitting description logic ontologies to abox and query examples,'' \emph{arXiv preprint arXiv:2508.08007}, 2025.

\bibitem{liu2024lost}
N.~F. Liu, K.~Lin, J.~Hewitt, A.~Paranjape, M.~Bevilacqua, F.~Petroni, and P.~Liang, ``Lost in the middle: How language models use long contexts,'' \emph{Trans. Assoc. Comput. Linguistics}, vol.~12, pp. 157--173, 2024.

\bibitem{yang2025qwen3}
A.~Yang, A.~Li, B.~Yang, B.~Zhang, B.~Hui, B.~Zheng, B.~Yu, C.~Gao, C.~Huang, C.~Lv \emph{et~al.}, ``Qwen3 technical report,'' \emph{arXiv preprint arXiv:2505.09388}, 2025, qwen~3.5-27B belongs to this model family.

\bibitem{silla2011hierarchical}
C.~N.~S. Jr. and A.~A. Freitas, ``A survey of hierarchical classification across different application domains,'' \emph{Data Mining and Knowledge Discovery}, vol.~22, no.~1, pp. 31--72, 2011.

\bibitem{sbert2019}
N.~Reimers and I.~Gurevych, ``{Sentence-BERT}: Sentence embeddings using siamese {BERT}-networks,'' in \emph{Proc. Conf. Empirical Methods in Natural Language Processing (EMNLP)}, 2019.

\end{thebibliography}

\end{document}